\documentclass{llncs}
\usepackage{times}
\usepackage{latexsym}
\usepackage{graphicx}
\usepackage{amsmath}
\usepackage{mathtools}
\usepackage{multirow}
\usepackage{booktabs}
\usepackage{array}

\usepackage{url}
\usepackage{makeidx}  
\urldef{\mailsa}\path|{binxuanh, kathleen.carley}@cs.cmu.edu, yanglano@andrew.cmu.edu|

\title{Aspect Level Sentiment Classification with Attention-over-Attention Neural Networks}

\author{Binxuan Huang, Yanglan Ou%
\and Kathleen M. Carley}
\institute{ Carnegie Mellon University,\\
5000 Forbe Ave., Pittsburgh, United States\\
\mailsa\\
}
\begin{document}
\maketitle
\begin{abstract}
Aspect-level sentiment classification aims to identify the sentiment expressed towards some aspects given context sentences. In this paper, we introduce an attention-over-attention (AOA) neural network for aspect level sentiment classification. Our approach models aspects and sentences in a joint way and explicitly captures the interaction between aspects and context sentences. With the AOA module, our model jointly learns the representations for aspects and sentences, and automatically focuses on the important parts in sentences. Our experiments on laptop and restaurant datasets demonstrate our approach outperforms previous LSTM-based architectures.
\end{abstract}

\section{Introduction}

Unlike document level sentiment classification task \cite{glorot2011domain,pang2002thumbs}, aspect level sentiment classification is a more fine-grained classification task. It aims at identifying the sentiment polarity (e.g. positive, negative, neutral) of one specific aspect in its context sentence. 
For example, given a sentence ``great food but the service was dreadful" the sentiment polarity for aspects ``food" and ``service" are positive and negative respectively. 

Aspect sentiment classification overcomes one limitation of document level sentiment classification when multiple aspects appear in one sentence. In our previous example, there are two aspects and the general sentiment of the whole sentence is mixed with positive and negative polarity. If we ignore the aspect information, it is hard to determine the polarity for a specified target. Such error commonly exists in the general sentiment classification tasks. In one recent work, Jiang et al. manually evaluated a Twitter sentiment classifier and showed that 40\% of sentiment classification errors are because of not considering targets \cite{jiang2011target}.


Many methods have been proposed to deal with aspect level sentiment classification. The typical way is to build a machine learning classifier by supervised training. Among these machine learning-based approaches, there are mainly two different types. One is to build a classifier based on manually created features \cite{jiang2011target,wagner2014dcu}. The other type is based on neural networks using end-to-end training without any prior knowledge \cite{ma2017interactive,tang2016aspect,wang2016attention}. Because of its capacity of learning representations from data without feature engineering, neural networks are becoming popular in this task.

Because of advantages of neural networks, we approach this aspect level sentiment classification problem based on long short-term memory (LSTM) neural networks. Previous LSTM-based methods mainly focus on modeling texts separately \cite{tang2015effective,wang2016attention}, while our approach models aspects and texts simultaneously using LSTMs.
Furthermore, the target representation and text representation generated from LSTMs interact with each other by an attention-over-attention (AOA) module \cite{cui2016attention}. AOA automatically generates mutual attentions not only from aspect-to-text but also text-to-aspect. This is inspired by the observation that only few words in a sentence contribute to the sentiment towards an aspect. Many times, those sentiment bearing words are highly correlated with the aspects. In our previous example, there are two aspects ``appetizers" and ``service" in the sentence ``the appetizers are ok, but the service is slow." Based on our language experience, we know that the negative word ``slow" is more likely to describe ``service" but not the ``appetizers". Similarly, for an aspect phrase, we also need to focus on the most important part. That is why we choose AOA to attend to the most important parts in both aspect and sentence. Compared to previous methods, our model 
performs better on the laptop and restaurant datasets from SemEval 2014 \cite{pontiki2016semeval}


\section{Related work}

\textbf{Sentiment Classification}\\
Sentiment classification aims at detecting the sentiment polarity for text. There are various approaches proposed for this research question \cite{medhat2014sentiment}. Most existing works use machine learning algorithms to classify texts in a supervision fashion. Algorithms like Naive Bayes and Support Vector Machine(SVM) are widely used in this problem \cite{liu2013scalable,pang2002thumbs,wang2012baselines}. The majority of these approaches either rely on n-gram features or manually designed features. Multiple sentiment lexicons are built for this purpose \cite{neviarouskaya2009sentiful,qiu2009expanding,taboada2011lexicon}.

In the recent years, sentiment classification has been advanced by neural networks significantly. Neural network based approaches automatically learn feature representations and do not require intensive feature engineering. Researchers proposed a variety of neural network architectures. Classical methods include Convolutional Neural Networks \cite{kim2014convolutional}, Recurrent Neural Networks \cite{lai2015recurrent,tang2015document}, Recursive Neural Networks \cite{socher2013recursive,zhu2015long}.  These approaches have achieved promising results on sentiment analysis. 

\noindent \textbf{Aspect Level Sentiment Classification}\\
Aspect level sentiment classification is a branch of sentiment classification, the goal of which is to identify the sentiment polarity of one specific aspect in a sentence. 
Some early works designed several rule based models for aspect level sentiment classification, such as  \cite{ding2007utility,nasukawa2003sentiment}. Nasukawa et al. first perform dependency parsing on sentences, then they use predefined rules to determine the sentiment about aspects \cite{nasukawa2003sentiment}. Jiang et al. improve the target-dependent sentiment classification by creating several target-dependent features based on the sentences' grammar structures \cite{jiang2011target}. These target-dependent features are further fed into an SVM classifier along with other content features.

Later, kinds of neural network based methods were introduced to solve this aspect level sentiment classification problem. Typical methods are based on LSTM neural networks. TD-LSTM approaches this problem by developing two LSTM networks to model the left and right contexts for an aspect target \cite{tang2015effective}. This method uses the last hidden states of these two LSTMs for predicting the sentiment. In order to better capture the important part in a sentence, Wang et al. use an aspect term embedding to generate an attention vector to concentrate on different parts of a sentence \cite{wang2016attention}. Along these lines, Ma et al. use two LSTM networks to model sentences and aspects separately \cite{ma2017interactive}. They further use the hidden states generated from sentences to calculate attentions to aspect targets by a pooling operation, and vice versa. Hence their IAN model can attend to both the important parts in sentences and targets. Their method is similar to ours. However, the pooling operation will ignore the interaction among word-pairs between sentences and targets, and experiments show our method is superior to their model.



\section{Method}

\noindent\textbf{Problem Definition}\\
In this aspect level sentiment classification problem, we are given a sentence $s=[w_1,w_2,...,w_i,..,w_j,...,w_n]$ and an aspect target $t=[w_i,w_{i+1},...,w_{i+m-1}]$. The aspect target could be a single word or a long phrase. The goal is to classify the sentiment polarity of the aspect target in the sentence. 

\begin{figure*}[!h]
    \centering
    \includegraphics[width=\textwidth]{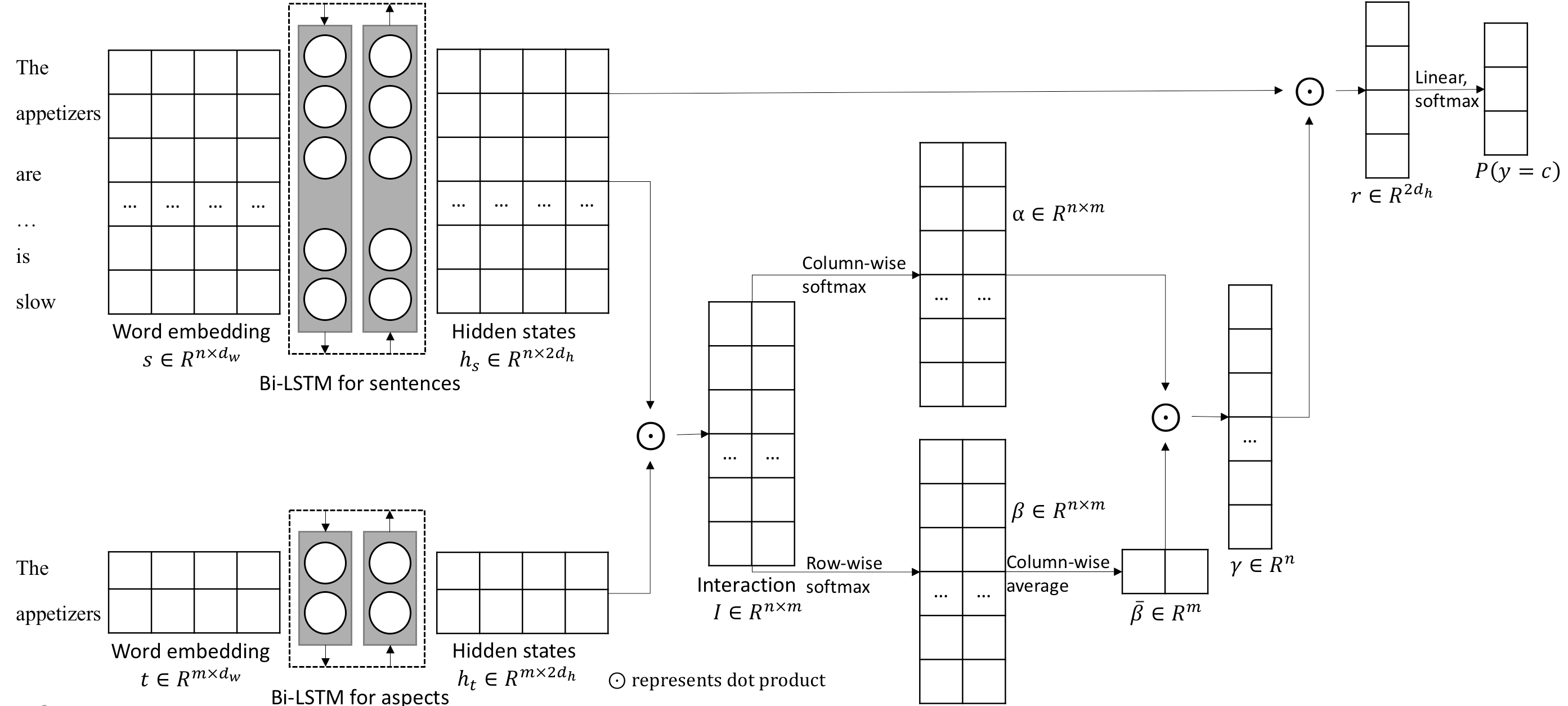}
     \vspace{-0.4cm}
    \caption{The overall architecture of our aspect level sentiment classification model.}
    \label{fig:arch}
\end{figure*}

The overall architecture of our neural model is shown in Figure \ref{fig:arch}. It is mainly composed of four components: word embedding, Bidirectional-Long short-term memory (Bi-LSTM), Attention-over-Attention module and the final prediction.

\noindent\textbf{Word Embedding}\\
Given a sentence $s=[w_1,w_2,...,w_i,..,w_j,$ $...,w_n]$ with length n and a target $t=[w_i,w_{i+1},...,w_{i+m-1}]$ with length m, we first map each word into a low-dimensional real-value vector, called word embedding \cite{bengio2003neural}. For each word $w_i$, we can get a vector $v_i\in R^{d_w}$ from $M^{V\times {d_w}}$, where $V$ is the vocabulary size and $d_w$ is the embedding dimension. After an embedding look up operation, we get two sets of word vectors $[v_1;v_2;...;v_n]\in R^{n\times d_w}$ and $[v_i;v_{i+1};...;v_{i+m-1}]\in R^{m\times d_w}$ for the sentence and aspect phrase respectively.

\noindent\textbf{Bi-LSTM}\\
After getting the word vectors, we feed these two sets of word vectors into two Bidirectional-LSTM networks respectively. We use these two Bi-LSTM networks to learn the hidden semantics of words in the sentence and the target. Each Bi-LSTM is obtained by stacking two LSTM networks. The advantage of using LSTM is that it can avoid the gradient vanishing or exploding problem and is good at learning long-term dependency \cite{hochreiter1997long}. 

With an input $s=[v_1;v_2;...;v_n]$ and a forward LSTM network , we generate a sequence of hidden states $\overrightarrow{h_s}\in R^{n\times d_h}$, where $d_h$ is the dimension of hidden states. We generate another state sequence $\overleftarrow{h_s}$ by feeding $s$ into another backward LSTM. In the Bi-LSTM network, the final output hidden states $h_s\in R^{n\times 2d_h}$ are generated by concatenating $\overrightarrow{h_s}$ and $\overleftarrow h_s$. We compute the hidden semantic states $h_t$ for the aspect target $t$ in the same way.
\begin{align}
\overrightarrow{h_s} & = \overrightarrow{LSTM}([v_1;v_2;...;v_n])\\
\overleftarrow{h_s} &= \overleftarrow{LSTM}([v_1;v_{2};...;v_n])\\
h_s &= [\overrightarrow{h_s}, \overleftarrow{h_s}]
\end{align}

\noindent\textbf{Attention-over-Attention}\\
Given the hidden semantic representations of the text and the aspect target generated by Bi-LSTMs, we calculate the attention weights for the text by an AOA module. This is inspired by the use of AOA in question answering \cite{cui2016attention}. Given the target representation $h_t\in R^{m\times 2d_h}$ and sentence representation $h_s\in R^{n\times 2d_h} $, we first calculate a pair-wise interaction matrix $I=h_s \cdot h_t^T$, where the value of each entry represents the correlation of a word pair among sentence and target. With a column-wise softmax and row-wise softmax, we get target-to-sentence attention $\alpha$ and sentence-to-target attention $\beta$. After column-wise averaging $\beta$, we get a target-level attention $\bar \beta \in R^m$, which indicating the important parts in an aspect target. The final sentence-level attention $\gamma\in R^n$ is calculated by a weighted sum of each individual target-to-sentence attention $\alpha$, given by equation (\ref{gamma}). By considering the contribution of each aspect word explicitly, we learn the important weights for each word in the sentence.
\begin{align}
\alpha_{ij} &= \frac{exp(I_{ij})}{\sum_i exp(I_{ij})}\\ 
\beta_{ij} &= \frac{exp(I_{ij})}{\sum_j exp(I_{ij})}\\
\bar \beta_j &= \frac{1}{n} \sum_{i} \beta_{ij}\\
\gamma &= \alpha \cdot \bar \beta^T
\label{gamma}
\end{align}

\noindent \textbf{Final Classification}\\
The final sentence representation is a weighted sum of sentence hidden semantic states using the sentence attention from AOA module.
\begin{align}
r = h_s^T \cdot \gamma
\end{align}

We regard this sentence representation as the final classification feature and feed it into a linear layer to project $r$ into the space of targeted $C$ classes.
\begin{align}
x = W_l \cdot r+b_l
\end{align}

\noindent where $W_l$ and $b_l$ are the weight matrix and bias respectively. Following the linear layer, we use a softmax layer to compute the probability of the sentence $s$ with sentiment polarity $c\in C$ towards an aspect $a$ as:
\begin{align}
P(y=c) = \frac{exp(x_c) }{\sum_{i\in C} exp(x_i)}
\end{align}

The final predicted sentiment polarity of an aspect target is just the label with the highest probability. We train our model to minimize the cross-entropy loss with $L_2$ regularization
\begin{align}
loss = -\sum_{i}\sum_{c\in C} I(y_i=c)\cdot log(P(y_i=c))+\lambda ||\theta||^2
\end{align}

\noindent where $I(\cdot)$ is an indicator function. $\lambda$ is the $L_2$ regularization parameter and $\theta$ is a set of weight matrices in LSTM networks and linear layer. We further apply dropout to avoid overfitting, where we randomly drop part of inputs of LSTM cells.

We use mini-batch stochastic gradient descent with Adam \cite{kingma2014adam} update rule to minimize the loss function with respect to the weight matrices and bias terms in our model.

\section{Experiments}
\noindent \textbf{Dataset}\\
We experiment on two domain-specific datasets for laptop and restaurant from SemEval 2014 Task 4 \cite{wagner2014dcu}. Experienced annotators tagged the aspect terms of the sentences and their polarities. Distribution by sentiment polarity category are given in Table \ref{data}.
\begin{table}[]
\centering
\begin{tabular}{|l|l|l|l|}
\hline
Dataset          & Positive & Neutral & Negative \\ \hline
Laptop-Train     & 994      & 464     & 870      \\ \hline
Laptop-Test      & 341      & 169     & 128      \\ \hline
Restaurant-Train & 2164     & 637     & 807      \\ \hline
Restaurant-Test  & 728      & 196     & 196      \\ \hline
\end{tabular}
\caption{Distribution by sentiment polarity category of the datasets from SemEval 2014 Task 4. Numbers in table represent numbers of sentence-aspect pairs.}
\label{data}
\end{table}
\vspace{-0.8cm}

\noindent \textbf{Hyperparameters Setting}\\
In experiments, we first randomly select 20\% of training data as validation set to tune the hyperparameters. All weight matrices are randomly initialized from uniform distribution $U(-10^{-4},10^{-4})$ and all bias terms are set to zero. The $L_2$ regularization coefficient is set to $10^{-4}$ and the dropout keep rate is set to 0.2\cite{srivastava2014dropout}. The word embeddings are initialized with 300-dimensional Glove vectors \cite{pennington2014glove} and are fixed during training. For the out of vocabulary words we initialize them randomly from uniform distribution $U(-0.01,0.01)$. The dimension of LSTM hidden states is set to 150. The initial learning rate is 0.01 for the Adam optimizer. If the training loss does not drop after every three epochs, we decrease the learning rate by half. The batch size is set as 25.

\noindent \textbf{Model Comparisons}\\
We train and evaluate our model on these two SemEval datasets separately. We use accuracy metric to measure the performance. In order to further validate the performance of our model, we compare it with several baseline methods. We list them as follows:

\textbf{Majority} is a basic baseline method, which assigns the largest sentiment polarity in the training set to each sample in the test set.

\textbf{LSTM} uses one LSTM network to model the sentence, and the last hidden state is used as the sentence representation for the final classification.

\textbf{TD-LSTM} uses two LSTM networks to model the preceding and following contexts surrounding the aspect term. The last hidden states of these two LSTM network are concatenated for predicting the sentiment polarity \cite{tang2015effective}.

\textbf{AT-LSTM} first models the sentence via a LSTM model. Then it combines the hidden states from the LSTM with the aspect term embedding to generate the attention vector. The final sentence representation is the weighted sum of the hidden states \cite{wang2016attention}.

\textbf{ATAE-LSTM} further extends AT-LSTM by appending the aspect embedding into each word vector \cite{wang2016attention}.

\textbf{IAN} uses two LSTM networks to model the sentence and aspect term respectively. It uses the hidden states from the sentence to generate an attention vector for the target, and vice versa. Based on these two attention vectors, it outputs a sentence representation and a target representation for classification \cite{ma2017interactive}. 
\textbf{\begin{table}[]
\centering
\begin{tabular}{|l|l|l|}
\hline
Methods   & Restaurant & Laptop \\ \hline
Majority  & 0.535      & 0.650  \\ \hline
LSTM      & 0.743      & 0.665  \\ \hline
TD-LSTM   \cite{tang2015effective} & 0.756      & 0.681  \\ \hline
AT-LSTM    \cite{wang2016attention}& 0.762      & 0.689  \\ \hline
ATAE-LSTM \cite{wang2016attention} & 0.772      & 0.687  \\ \hline
IAN      \cite{ma2017interactive}  & 0.786      & 0.721  \\ \hline
AOA-LSTM         &    \textbf{0.812} (0.797$\pm$0.008)        &  \textbf{0.745} (0.726$\pm$0.008)      \\ \hline
\end{tabular}
\caption{Comparison results. For our method, we run it 10 times and show "best (mean$\pm$std)". Performance of baselines are cited from their original papers.
}\label{compare}
\end{table}}
\vspace{-0.8cm}

In our implementation, we found that the performance fluctuates with different random initialization, which is a well-known issue in training neural networks \cite{sutskever2013importance}. Hence, we ran our training algorithms 10 times, and report the average accuracy as well as the best one we got in Table \ref{compare}. All the baseline methods only reported a single best number in their papers. On average, our algorithm is better than these baseline methods and our best trained model outperforms them in a large margin. 

\noindent \textbf{Case Study}\\
In Table \ref{case}, We list five examples from the test set. To analyze which word contributes the most to the aspect sentiment polarity, we visualize the final sentence attention vectors $\gamma$ in Table \ref{case}. The color depth indicates the importance of a word in a sentence, the darker the more important. In the first two examples, there are two aspects ``appetizers" and ``service" in the sentence ``the appetizers are ok, but the service is slow." We can observe that when there are two aspects in the sentence, our model can automatically point to the right sentiment indicating words for each aspect. Same thing also happens in the third and fourth examples. In the last example, the aspect is a phrase ``boot time." From the sentence content ``boot time is super fast, around any where from 35 seconds to 1 minute," this model can learn ``time" is the most important word in the aspect, which further helps it find out the sentiment indicating part ``super fast."

\begin{table*}[!h]
\centering
\begin{tabular}{|c|c|c|c}
\cline{1-3}
Aspect                                                                                            & Sentence                                                                                         & Ans./Pred. & \multirow{6}{*}{ \begin{minipage}{.08\textwidth} \ \\ \ \\ \ \\ \includegraphics[width=\linewidth]{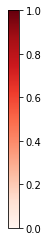}\end{minipage}} \\ \cline{1-3}
appetizers                                                                                        & \begin{minipage}{0.6\textwidth} \vspace{0.05cm} 
\includegraphics[width=\linewidth]{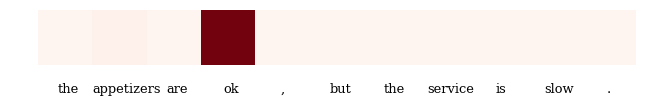}\end{minipage}  & 0/0               &                                                                                                                   \\ \cline{1-3}
service                                                                                           & \begin{minipage}{.6\textwidth}\vspace{0.05cm} \includegraphics[width=\linewidth]{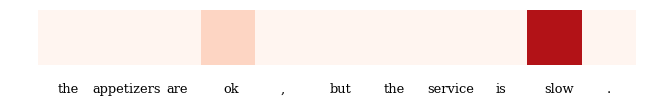}\end{minipage}  & -1/-1             &                                                                                                                   \\ \cline{1-3}
food                                                                                              & \begin{minipage}{.6\textwidth}\vspace{0.05cm}\includegraphics[width=\linewidth]{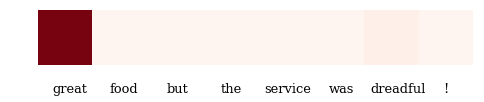}\end{minipage}  & +1/+1             &                                                                                                                   \\ \cline{1-3}
service                                                                                           & \begin{minipage}{.6\textwidth}\vspace{0.05cm}\includegraphics[width=\linewidth]{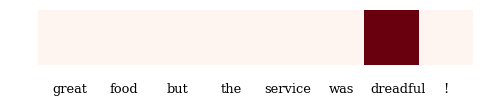}\end{minipage}  & -1/-1             &                                                                                                                   \\ \cline{1-3}
\begin{minipage}{.13\textwidth}\vspace{0.05cm} \includegraphics[width=\linewidth]{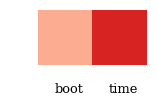}\end{minipage} & \begin{minipage}{.6\textwidth}\includegraphics[width=\linewidth]{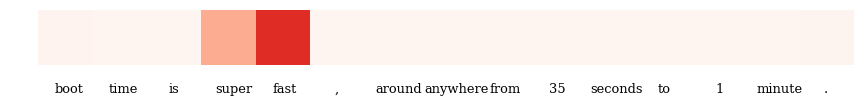}\end{minipage} & +1/+1             &                                                                                                                   \\ \cline{1-3}
\end{tabular}
  \caption{Examples of final attention weights for sentences. The color depth denotes the importance degree of the weight in attention vector $\gamma$. }
    \label{case}
\end{table*}

\vspace{-0.8cm}


\noindent \textbf{Error Analysis}\\
The first type of major errors comes from non-compositional sentiment expression which also appears in previous works \cite{tang2016aspect}.  For example, in the sentence ``it took about 2 1/2 hours to be served our 2 courses," there is no direct sentiment expressed towards the aspect ``served." Second type of errors is caused by idioms used in the sentences. Examples include ``the service was on point - what else you would expect from a ritz?" where ``service" is the aspect word. In this case, our model cannot understand the sentiment expressed by idiom ``on point." The third factor is complex sentiment expression like ``i have never had a bad meal (or bad service) @ pigalle." Our model still misunderstands the meaning this complex expressions, even though it can handle simple negation like ``definitely not edible" in sentence ``when the dish arrived it was blazing with green chillis, definitely not edible by a human".

\section{Conclusion}
In this paper, we propose a neural network model for aspect level sentiment classification. 
Our model utilizes an Attention-over-Attention module to learn the important parts in the aspect and sentence, which generates the final representation of the sentence. Experiments on SemEval 2014 datasets show superior performance of our model when compared to those baseline methods. Our case study also shows that our model learns the important parts in the sentence as well as in the target effectively.

In our error analysis, there are cases that our model cannot handle efficiently. One is the complex sentiment expression. One possible solution is to incorporate sentences' grammar structures into the classification model. Another type of error comes from uncommon idioms. In future work, we would like to explore how to combine prior language knowledge into such neural network models.

\bibliographystyle{splncs03.bst}
\bibliography{reference}

\end{document}